\documentclass[11pt]{article}
\usepackage{vocab}
\usepackage{times}
\usepackage{url}
\usepackage{latexsym}
\usepackage{graphicx}
\usepackage[export]{adjustbox}
\usepackage{subfig}
\usepackage[most]{tcolorbox}
\usepackage{lipsum}
\usepackage{hyperref}

\let\citet\newcite
\let\citep\cite

\graphicspath{ {./images/} }

\colingfinalcopy 

\title{Employing distributional semantics to organize task-focused
  \mbox{vocabulary learning}}

\author{\begin{tabular}{cc}
  Haemanth Santhi Ponnusamy & Detmar Meurers\\
        \end{tabular}\\
\rule{0em}{3ex}University of Tübingen, Germany\\
{\tt \{hsp,dm\}@sfs.uni-tuebingen.de}\\}

\date{}

\begin{document}
\maketitle
\begin{abstract}
  How can a learner systematically prepare for reading a book they are
  interested in? In this paper, we explore how computational
  linguistic methods such as distributional semantics, morphological
  clustering, and exercise generation can be combined with graph-based
  learner models to answer this question both conceptually and in
  practice. Based on the highly structured learner model and concepts
  from network analysis, the learner is guided to efficiently explore
  the targeted lexical space. They practice using multi-gap learning
  activities generated from the book focused on words that are central
  to the targeted lexical space. As such the approach offers a unique
  combination of computational linguistic methods with concepts from
  network analysis and the tutoring system domain to support learners
  in achieving their individual, reading task-based learning goals.
\end{abstract}

\section{Introduction}
\label{sec:intro}

Learning vocabulary is a major component of foreign language learning.
In the school context, initially vocabulary learning is typically
organized around the words introduced by the text book. In addition to
the incrementally growing vocabulary lists, some textbooks also
provide thematically organized word banks. When other texts are read,
the publisher or the teacher often provides annotations for new
vocabulary items that appear in the text.  A wide range of digital
tools have been developed to support such vocabulary learning, from
digital versions of file cards to digital text editions offering
annotations.

While such applications serve the needs of the formal learning setting
in the initial foreign language learning phase, where the texts that
are read are primarily chosen to systematically introduce the
language, later the selection of texts to be read can in principle
follow the individual interests of the student or adult, which boosts
the motivation to engage with the book. Linking language learning to a
functional goal that someone actually wants to achieve using language
is in line with the idea of Task-Based Language Teaching (TBLT) as a
prominent strand of foreign language education
\citep{Willis.Willis-13}.

Naturally, not all authentic texts are accessible to every learner,
but linguistically-aware search engines, such as FLAIR
\cite{Chinkina.Meurers-16}, make it possible to identify authentic
texts that are at the right reading level and are rich in the language
constructions next on the curriculum. Where the unknown vocabulary
that the reader encounters in such a setting goes beyond the around
2\% of unknown words in a text that can be present without substantial
loss of comprehension \citep{Schmitt.Jiang.Grabe-11}, many digital
reading environments provide the option to look up a word in a
dictionary. Yet, frequently looking up words in such a context is
cumbersome and distracts the reader from the world of the book they
are trying to engage with. Relatedly, one of the key criteria of TBLT
is that learners should rely on their own resources to complete a task
\cite{Ellis-09b}. But this naturally can require pre-task activities
preparing the learner to be able to successfully tackle the task
\citep{Willis.Willis-13}. But how can a learner systematically prepare
for reading a text or book they are interested in reading?

In this paper, we explore how computational linguistic methods such as
distributional semantics, morphological clustering, and exercise
generation can be combined with graph-based learner models to answer
this question both conceptually and in practice. On the practical
side, we developed an application that supports vocabulary learning as
a pre-task activity for reading a self-selected book. The conceptual
goal is to automatically organize the lexical semantic space of any
given English book in the form of a graph that makes it possible to
sequence the vocabulary learning in a way efficiently exploring the
space and to visualize this graph for the users as an open learner
model \citep{Bull.Kay-10} showing their growing mastery of the book's
lexical space.  Lexical learning is fostered and monitored through
automatically generated multi-gap activities
\cite{zesch-melamud-2014-automatic} that support learning and revision
of words in the contexts in which they occur in the book.

In section~\ref{sec:graph} we discuss how a book or other text chosen
by the learner is turned in to a graph encoding the lexical space that
the learner needs to engage with to read the book, and how words that
are morphologically related as word families \cite{bauer1993word} are
automatically identified and compactly represented in the graph
(\ref{sec:word-families}). In section~\ref{sec:modeling-knowledge} we
then turn to the use of the graph representation of the lexical
semantic space of the book to determine the reader's learning path and
represent their growing lexical knowledge as spreading activation in
the graph. In section~\ref{sec:application}, the conceptual ideas are
realized in an application. We discuss how the new learner cold-start
problem is avoided using a very quick word recognition task we
implemented, before discussing the content selection and activity
generation for practice and testing activities.
Section~\ref{sec:evaluation-and-related-work} then provides a
conceptual evaluation of the approach and compares it with related
with, before wrapping up with a conclusion in
section~\ref{sec:conclusion}.





\section{Constructing a graph for the lexical space of a book: a
  structured domain model}
\label{sec:graph}

Going beyond the benefits of interactivity and adaptivity of
individualized digital learning tools, supporting learner autonomy is
known to be important for boosting motivation and self-regulation
skills \cite{Godwin-Jones-19}. This includes the choice of reading
material a learner wants to engage with, where the texts prepared by a
teacher or publisher cannot reflect the interests of individual
students, the topics and genres they want to explore in the foreign
language. The freedom of choosing a text that the learner wants to
engage with also identifying a clear functional goal for learning
vocabulary -- learning new words to enable us to read a text of
interest, so that the interest in the content coincides with the
interest in further developing the language skills.  In that sense
learning vocabulary becomes a pre-task activity in the spirit of
task-based language learning. Organizing vocabulary learning in this
way also helps turn the otherwise open-ended challenge of learning the
lexical space of a new language to the clearly delineated task of
mastering a sub-space. This functionally guided approach contrasts
with the approach of other vocabulary learning tools selecting random
infrequent lexical items from the language to be learned, which given
their rare and often highly specialized nature are likely to only be
useful for impressing friends when playing foreign language scrabble.

To make text-driven vocabulary learning work, we need to map the text
selected by the learner into a structured domain to support systematic
and efficient learning of the lexical space as used in the book. We
distinguish the process of structuring the vocabulary used in the
book, independent of the learner's background, from the representation
of the individual learner's knowledge. The former is tackled in this
section and can be regarded as our domain model, while the latter is a
learner model that essentially is an overlay over the domain model,
and will be discussed in section~\ref{sec:modeling-knowledge}.

Since vocabulary learning is about establishing form-meaning
connections, in principle the basic unit best suited for this would be
word senses. At the same time, full automatic word sense
disambiguation is complex, error prone, and often domain specific --
and in the context of a given book, a given word will often occur with
the same meaning. We therefore limit ourselves to only disambiguating
homographs in terms of their part-of-speech, following
\newcite{wilks1998grammar}. Throughout our approach, we therefore use
$<$word, POS$>$ pairs as basic units. To POS annotate the book
selected by the user, we use the Spacy NLP tools
(\url{http://spacy.io}). Given our focus on learning the
characteristic vocabulary of the book, we eliminate stop
words 
as well as word-POS pairs appearing less than five times in the given book.

\subsection{Semantic and thematic relations}
\label{sec:semantic-thematic}

To structure the lexical space in terms of meaning, there are two
related options. Words can be \emph{semantically related}, for
example, \textit{tiger}, \textit{lion}, \textit{elephant}, and
\textit{crocodile} all have the property of being wild animals, on
from the perspective of a WordNet, they are hyponyms of \textit{wild
  animal}. On the other hand, words can also be \emph{thematically
  related}, such as \textit{blackboard}, \textit{teacher}, and
\textit{chalk} all belonging to a school theme. 
\newcite{gholami2014semantic} highlights the benefits of the semantic
approach over the thematic approach from the perspective of a
tutor. As we are building a system that acts like a tutor tracking and
fostering the learner's vocabulary knowledge, we decided to focus on
semantic relatedness.

\subsubsection{Word families}
\label{sec:word-families}

\begin{figure}[htbp]
\begin{center}
\scalebox{.8}{\begin{tabular}{@{}llll@{}}
2 & \begin{tabular}[c]{@{}l@{}}develop\\ develops\\ developed\\ developing\end{tabular}              & \begin{tabular}[c]{@{}l@{}}wood\\ wood's\\ woods\\ wooded\end{tabular}         & \begin{tabular}[c]{@{}l@{}}bright\\ brighter\\ brightest\end{tabular}     \\ \hline
3 & \begin{tabular}[c]{@{}l@{}}developable\\ undevelopable\\ developers(s)\\ undeveloped\end{tabular} & \begin{tabular}[c]{@{}l@{}}woody\\ woodiest\\ woodier\\ woodiness\end{tabular} & \begin{tabular}[c]{@{}l@{}}brightly\\ brightish\\ brightness\end{tabular} \\ \hline
4 & \begin{tabular}[c]{@{}l@{}}development(s)\\ developmental\\ developmentally\end{tabular}         & \textbf{}                                                                               & \textbf{}                                                                          \\ \hline
5 & \begin{tabular}[c]{@{}l@{}}developmentwise\\ semideveloped\\ antidevelopment\end{tabular}        & wooden                                                                         & brighten                                                                  \\ \hline
6 & \begin{tabular}[c]{@{}l@{}}redeveloped\\ predevelopment\end{tabular}                              & anti-wooden                                                                    & \textbf{}                                                                          \\ \hline
              \end{tabular}}\hspace{6em}\includegraphics[scale=0.26]{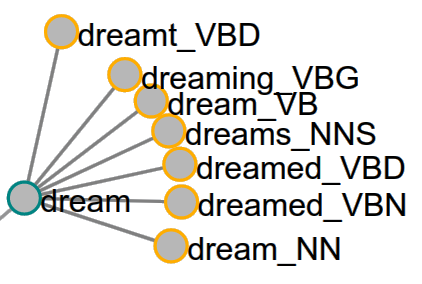}
\end{center}
\caption{A word family example \protect\citep{bauer1993word} and an
  expanded family node in our graph}
\label{wordfamily-table} 
\end{figure}

Complementing the lexical semantic relationships, words are also
related to each other through derivational and inflectional
morphology.  Many of these morphological processes are semantically
transparent.  \newcite{bauer1993word} proposed the idea of grouping
words into so-called \emph{word families} stating that ``once the base
word or even a derived word is known, the recognition of other members
of the family requires little or no extra effort''. The creation of
word families is based on criteria involving frequency, regularity,
productivity and predictability of all the English affixes.
\newcite{bauer1993word} arranged the inflectional affixes and common
derivational affixes into the graded levels, as exemplified on the
left-hand side of Figure \ref{wordfamily-table}.


We adopt the idea of word families to compactly represent
morphologically related words.  The graph on the right side of
Figure~\ref{wordfamily-table} exemplifies the word family that becomes
visible when selecting the lemma \emph{dream} in our graph
representation (where word families normally are shown in collapsed
form and represented by their underlying lemma).  We currently
collapse words belonging to any of the levels into one word family,
though in the future one could chose to collapse only those levels
that are semantically most transparently connected.

%
%

\subsection{Generating a lexical graph of word families and their
  semantic relations}


To structure the lexical space of the user selected book in terms of a
semantically related word graph, we start with a distributional
semantic vector representation of each word, which we obtain from the
pre-trained model of GloVe \cite{pennington2014glove} based on the
co-occurrence statistics of the the words form a large Common Crawl
data-set (\url{http://commoncrawl.org}).

\begin{figure}[htbp!]
  \centering
  \parbox{.35\textwidth}{$ w_{12} = \max_{i \in F_1; j \in F_2}
    \frac{V_i * V_j}{\|V_i\| \|V_j\|}
    $\\[2ex]
    where, ${w_{12}}$ is the cosine similarity between the families
    ${F_1}$ and ${F_2}$}
  \vspace{-1ex}\includegraphics[width=.38\textwidth,valign=M]{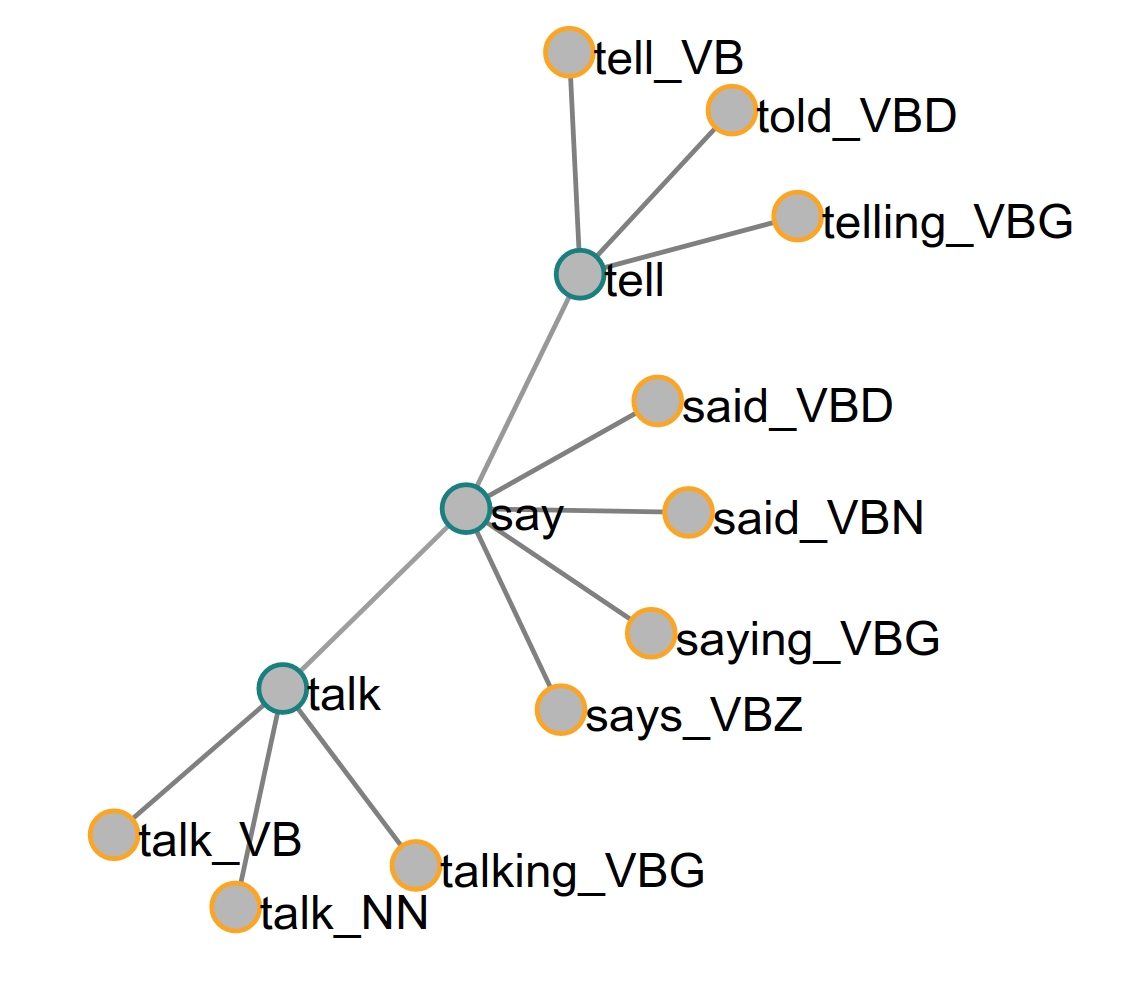}
\caption{Formula for computing relationship between families and a resulting example}
\label{fig:network}
\end{figure}

On this basis, the relationship score between the families are
computed to be the maximum of pair-wise cosine similarity score of all
its members. Let ${F_1}$ be a family with ${m}$ members and ${F_2}$ be
a family with ${n}$ members. The relationship score between the two
family ${F_1}$ and ${F_2}$ is the maximum of cosine similarity score
of all ${m \times n}$ pairs, as spelled out in Figure~\ref{fig:network}.
%
%
In this way, we create a fully connected network of word families, as
exemplified in the figure. The families with members that are closer
in the semantic vector space are connected with higher weights.

Following \newcite{d1997mathematical}, the number of edges in the
graph can be computed as \mbox{$e = \frac{ n \times (n - 1)}{2}$}, where $e$
is the total number of edges and $n$ is the number of nodes or
families in the graph.  The number of edges in the graph thus grows
exponentially as the number of node increases.

When inspecting graphs derived for sample texts, we observe the
majority of the connections are weak. To obtain a graph of semantic
relationships that meaningfully structure the vocabulary used in a
book, we focus on stronger relationship and eliminate edges with
weights less than 0.3.

We also observed that the node families of very frequently occurring
verbs tend to be very densely connected, and this impact of frequency
on distributional semantic measures has been discussed in the
literature \cite{patel1998extracting,weeds2004characterising}. In
order to control for this kind of over-sensitivity of distributional
semantic measures for highly frequent words, we restrict the node
degree to a maximum threshold, which based on experiments with sample
data we set to five. So for each node, only the five edges with the
highest weight are retained.

As a result of the method described in this section, we obtain a
lexical graph for the user-provided text that structures and compactly
represents the lexical space of the text in a graph-based domain
model. This is the lexical space that the user wants to explore and
master enough to be able to read the book. In terms of computational
linguistic methods, on the one hand, distributional semantics creates
the overall structure of a \textbf{meaning}-connected lexical space,
on the hand, word families organize and collapse \textbf{forms} that
are related by morphological processes in the linguistic system.

\subsection{Example generation of graphs for books}

\begin{figure}[htbp]
   \centering
   \vspace*{-2ex}
    \includegraphics[width=.35\linewidth, trim={0 1cm 0 1cm},clip = true]{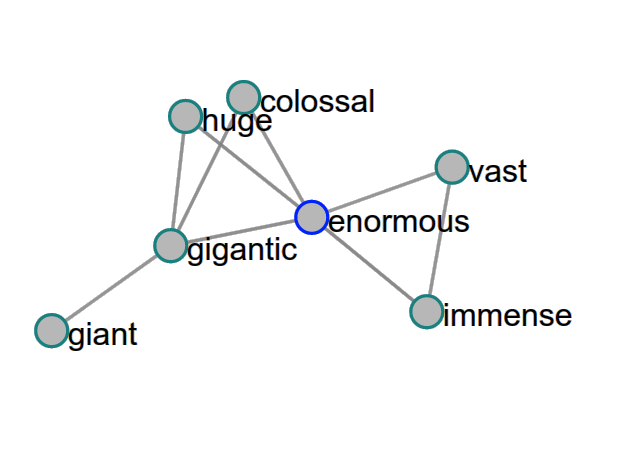}\qquad
    \includegraphics[width=.3\linewidth, trim={0 1cm 0 1cm},clip = true]{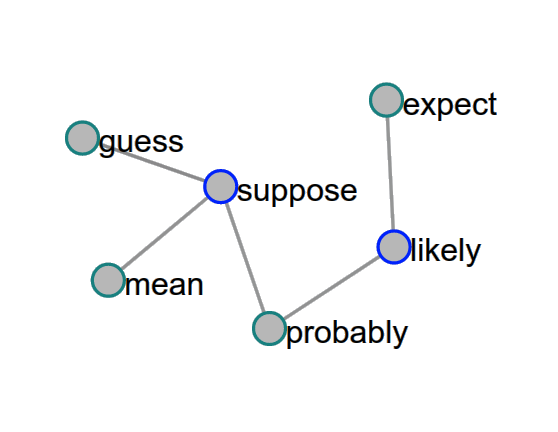}\\[-4ex]
    \includegraphics[width=.45\linewidth, trim={0 0 0 1cm},clip = true]{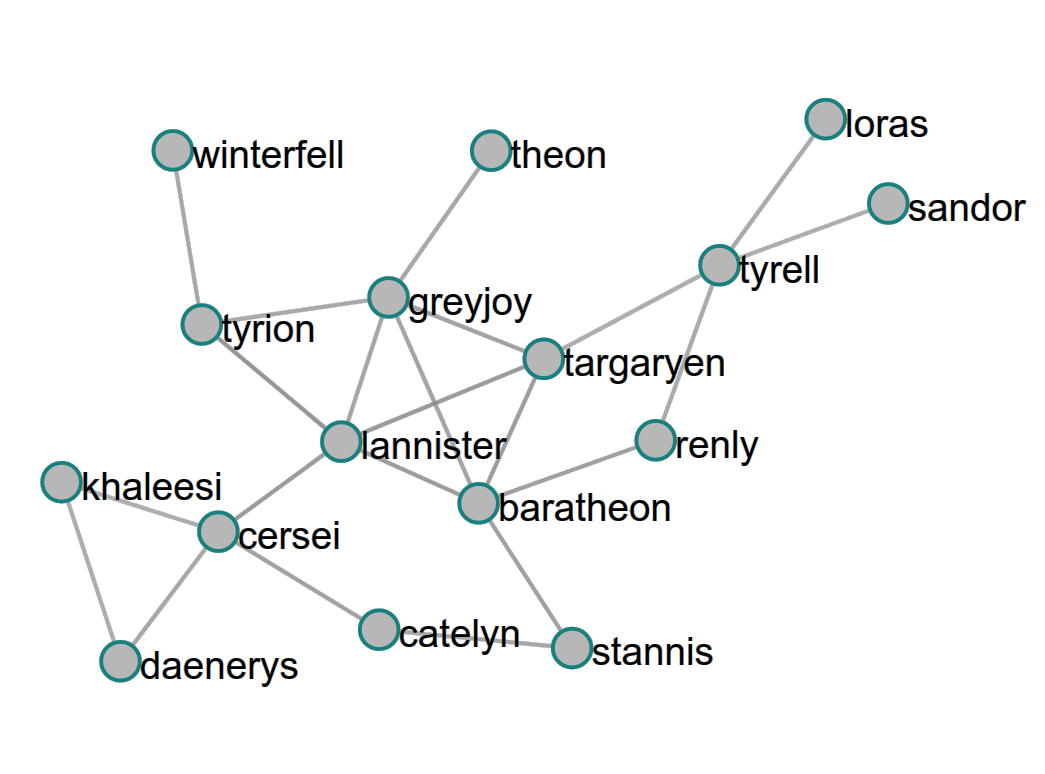}\qquad
    \includegraphics[width=.35\linewidth, trim={0 0 0 1cm},clip = true]{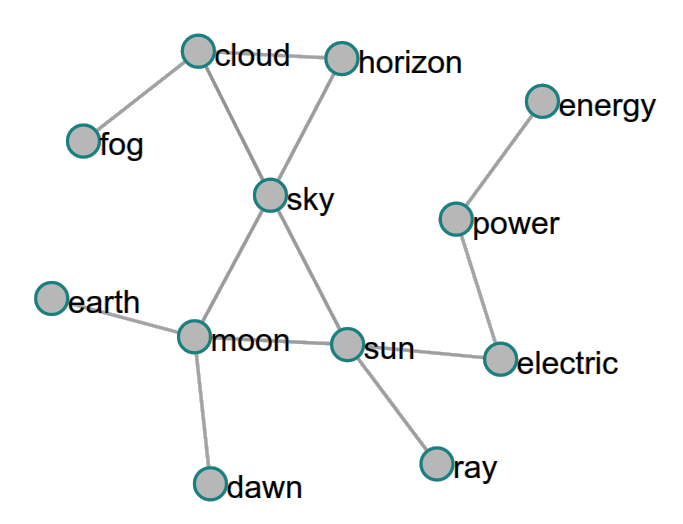}\vspace{-1ex}
  \caption{Example family clusters from the graphs resulting for the sample books}
  \label{fig:clusters}
\end{figure}

To test the graph construction, we chose three books as a sample to
study the characteristics of the vocabulary space created by our
application: (a) Twenty Thousand Leagues Under the Sea by Jules Verne,
(b) Harry Potter and the Sorcerer's Stone by J. K. Rowling and (c) A
Game of Thrones: A Song of Ice and Fire by George R. R. Martin.  Table
\ref{bookstats-table}
\begin{table}[htb]
\begin{tabular}{ p{6.8cm} p{1.3cm} p{1.5cm} p{1.3cm}  p{1.3cm} p{1.3cm} }
\textbf{Book title}                     & \textbf{Unique words} & \textbf{Learning targets} & \textbf{Graph nodes} & \textbf{Graph edges} & \textbf{Graph size} \\ \hline 
Twenty Thousand Leagues Under the Sea    & 10k                                        & 1.7k                                          & 1.3k                                                 & 3.3k                                                         & 18 MB                                             \\
Harry Potter and the Sorcerer's Stone   & 6.5k                                       & 1.2k                                          & 1k                                                   & 2.4k                                                         & 13 MB                                              \\
A Game of Thrones: A Song of Ice and Fire & 14k                                        & 3.7k                                          & 2.5k                                                 & 5.6k                                                         & 35 MB                                               \\
\end{tabular}

\caption{Characteristics of the graphs derived for vocabulary space for three books}
\label{bookstats-table} 
\end{table}
indicates the size of the text and the graph created for each book. We
see that 15--25\% of the words from the text qualifies as the lexical
learning targets.  20--30\% of those collapse into a family encoded by
a graph node.  The average connections a family has with other
families is around 2.5. The algorithm thus seems to work as expected
for these sample books.  Some example word family clusters formed for
these books at a threshold of similarity scores greater than 0.7 are
shown in Figure \ref{fig:clusters}, with only the root nodes of each
family being displayed here.



\section{Representing the lexical knowledge of a learner: an open
  learner model}
\label{sec:modeling-knowledge}

With a structured domain model established for the vocabulary space to
be explored by the user, we want to make use of it to efficiently
guide the learner to cover the space and track learning in a learner
model. The learner model is an overlay on the domain model that helps
us track the learner's vocabulary knowledge in terms of a mastery
score associated with each word family.
On the basis of the learner model, we then can propose the next set of
words to be practiced in a way that reduces the number of interaction
required to cover the vocabulary space. It also serves as an open
learner model \cite{Bull.Kay-10} by allowing the user to view and
explore the lexical space of the book as a graph, with each node being
colored according to the current mastery score. In this section we
discuss how this is achieved.

\subsection{Central node selection for efficient  exploration of the vocabulary graph}
\label{sec:central-node-selection}

Identifying the nodes that are more central than others is one of the
vital task in network analysis
\cite{freeman1978centrality,bonacich1987power,borgatti2005centrality,borgatti2006graph}. \newcite{freeman1978centrality}
formulated three major centrality measures for a node in a network:
(a) \emph{degree centrality:} Measure of strength of ties of each
nodes in the network, (b) \emph{closeness centrality:} Measure of
closeness of a node to all other nodes in the network, and (c)
\emph{betweenness centrality}: Measure of cardinality (number of
elements) of a set S, the set of shortest paths of other node pairs in
the network that passes through a node. 

The degree centrality measure is a greedy approach looking only at the
immediate neighbours to decide the central node, whereas the closeness
centrality measure accounts for the bigger picture of the entire
network. So closeness centrality seems best suited for our goal of
efficient coverage of the network, in our case: the graph representing
the vocabulary of the given book.  However, closeness centrality
cannot as such be applied to networks with disconnected components --
which is problematic since we may well obtain networks with many
disconnected components due to the pruning of weak links discussed in
the previous section. Fortunately, \newcite{wasserman1994social}
proposes an improved formula that also works for such graphs.
Based on this metric, we choose the top 20 words for a learning
session.

\begin{figure}[htb]
\centering
\includegraphics[width=\textwidth, clip,trim=50 0 400 200]{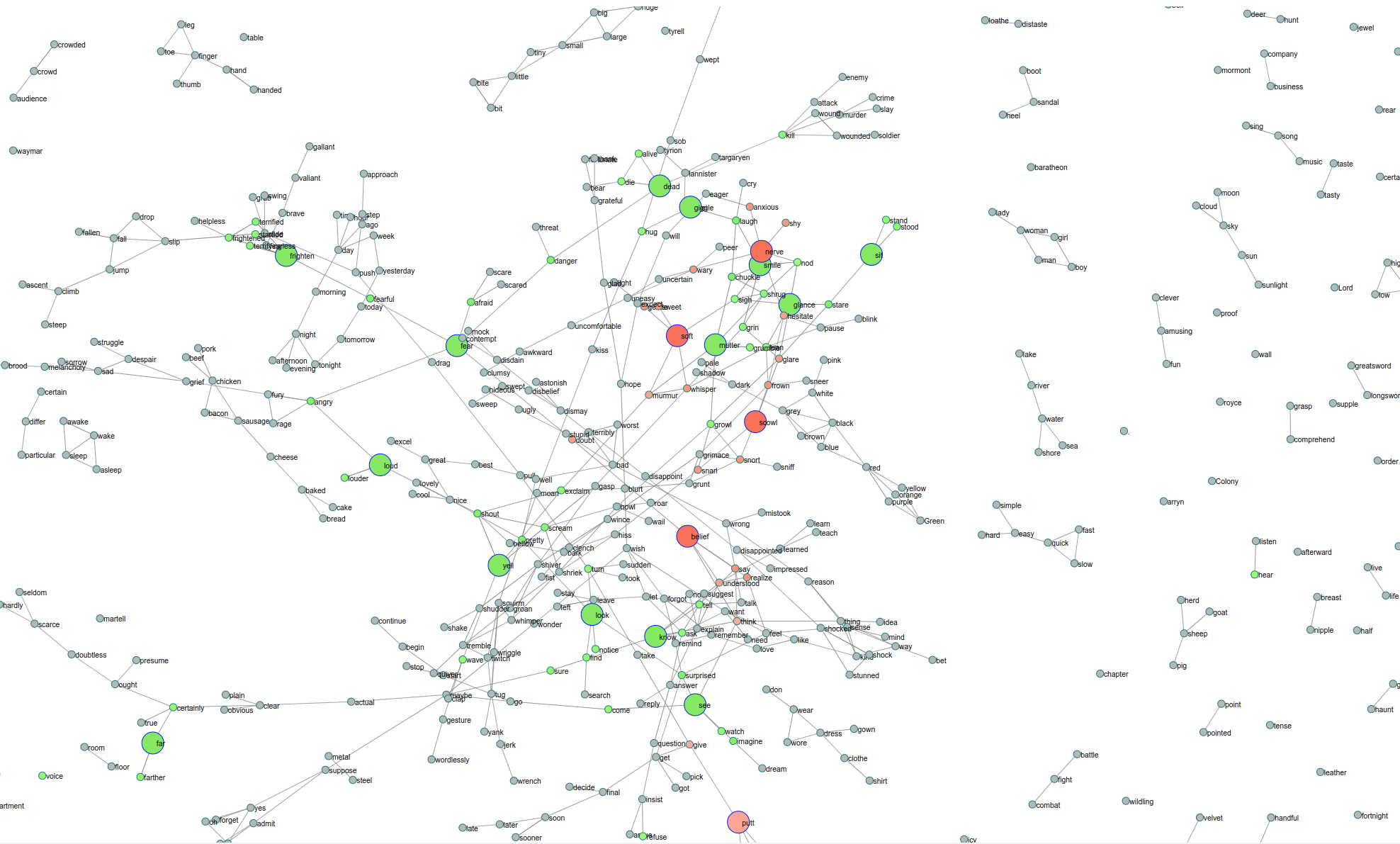}
\caption{A lexical graph for a book with highlighted central nodes
  selected for the next learning session}
\label{fig:graph-overview}
\end{figure}

Selecting the next words to be learned based on closeness centrality
brings up the problem that neighbors that are tightly bound to the
central node are likely have a similar closeness centrality score. So
when selecting the words to be practiced only based on closeness
centrality, we would risk practicing closely related lexical items
rather than systematically introducing the learner to the broader
lexical space. In order to avoid this issue, we exclude the immediate
neighbours of a word that was selected from that learning session.
The resulting approach supports a more distributed selection of words,
as illustrated by Figure \ref{fig:graph-overview} showing an example
with highlighted central nodes selected for the next learning session.

\subsection{Mastery scores and updating them in the graph to capture learning}
\label{sec:mastery-scores}

Each node in the graph is associated with a mastery score ranging from
0 to 1, with 1 indicating that the learner masters the word. We
initialize the master score of each node with 0.5 and interpret this
as a middle ground, where the model is uncertain about the learner's
knowledge about that word.

The mastery score is updated based on the learner responses in the
learning activities. To address this bottleneck that the system is
tied to such a thin stream of evidence about the learner's lexical
knowledge, we make use of the fact that the learner model is based on
a network of semantically related word families.  We use this to 
spread some activation from a word where the learner has shown mastery
to semantically closely related words to indicate that this word is
more likely to also be known.

Let $r$ be the learner response for a learning activity involving a
word from the family ${F_i}$. Then the update to its mastery ${m_i}$
is updated using $ \Delta m_i = m_i * \alpha * r$.  The update to the
mastery score of its immediate neighbours is weighted based on the
similarity score between the families
$ \Delta m_j = m_j * (\beta * r * w_{ij}) $ where ${m_j}$ is the
mastery score of ${F_j}$, a neighbouring family of ${F_i}$ attached
with a edge weight of ${w_{ij}}$. ${\alpha}$ and ${\beta}$ are
tune-able parameters for the magnitude of an update.
${r \in \{-1, +1\}}$ indicate the polarity of the learner's response,
$+1$ for the learner responding correctly and $-1$ an incorrect
response.

\begin{figure}[htb]
\centering\vspace{-1ex}\includegraphics[width=.53\textwidth,clip,trim=12 0 0 2]{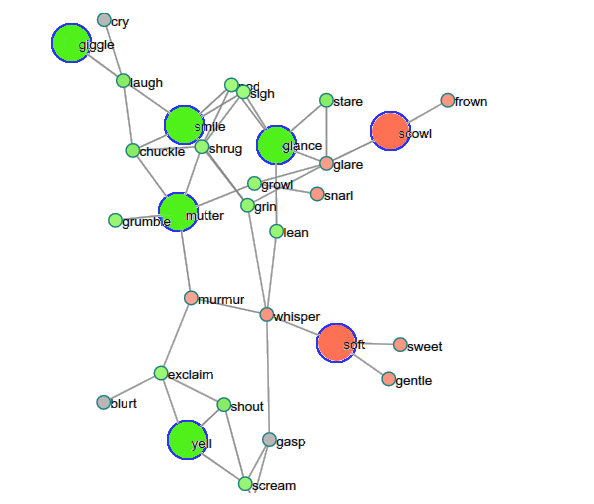}\hspace{-4cm}
\includegraphics[width=.6\textwidth,clip,trim=2 0 2 5]{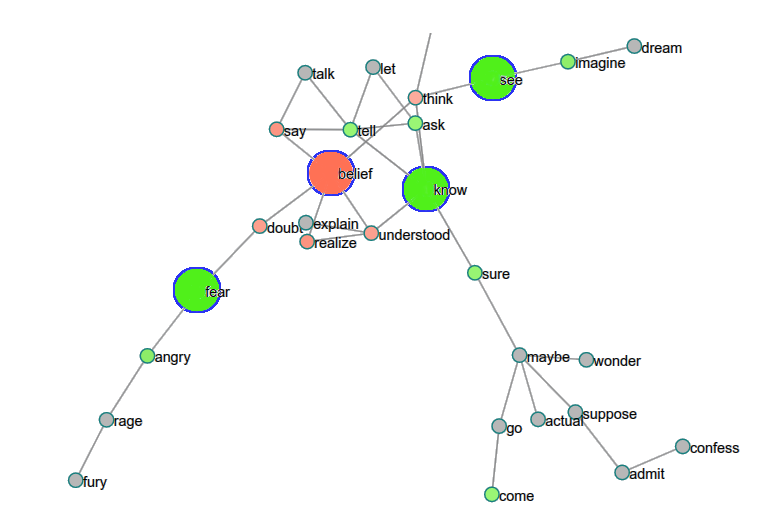}\vspace{-1ex}
\caption{A close-up view of a learner model showing nodes selected for
  practice}
\label{fig:graph2}
\end{figure}

Figure \ref{fig:graph2} provides a close-up view of the graph with
enlarged nodes highlighting the nodes selected for a learning
activities.
%
%
The figure also illustrates the color representation of the mastery
level and the spreading activation to neighboring nodes.  Initially
all nodes are grey, corresponding to mastery level of 0.5 that has not
yet been touched during learning. The closer the level gets to 1, the
more green the node appears, and the closer to 0, the more red. A node
the user has sometimes shown mastery for and sometimes not so that it
returns to the 0.5 level is shown in yellow.

While the colored visualization of the lexical space serves as an open
learner model allowing learners to inspect the current state of their
knowledge in relation to the lexical demands of the book, the mastery
level also plays a role in the selection of the next words to be
practiced.  Words with a mastery level over 0.8 are no longer
selected. In the future, we plan to add a component that takes into
account memory decay and the so-called spacing effect
\citep{Sense.Behrens.ea-16} to optimize when a word is selected again.

\section{Putting it all together in an application}
\label{sec:application}

\subsection{A warm start for the learner model}

Given that we are targeting learners beyond the beginner stage, it is
important to determine their vocabulary knowledge to avoid a cold
start of the learner model. We do not want to start from a blank slate
since that requires many interactions with the system until the
learner model reflects the learner's lexical competence -- a time
during which the system cannot optimally select the words to be
learned next.

To avoid this cold start problem, we implemented a short web-based
vocabulary Yes/No test, which has long been used for vocabulary
estimation
\cite{sims1929reliability,tilley1936technique,goulden1990large}.The
participants are provided with a checklist of words and the user has
to select whether they know the word or not. While there is a rich
literature on the test and various adjustments have been proposed to
counter its weaknesses as a competence diagnostic
\cite{meara1987alternative,beeckmans2001examining,huibregtse2002scores,mochida2006yes},
for our goal of allowing learners to quickly initialize their learner
model, it is very well suited.

The words included in the test are selected from the graph using the
same central node selection approach we introduced in
section~\ref{sec:central-node-selection}, and the mechanism spreading
activation to related nodes discussed in
section~\ref{sec:mastery-scores} allows the system to make additional
use of the information from the fast initial test.





\subsection{Activity generation: practicing and testing in the target context}

Content creation for vocabulary learning activities typically is a
task requiring human effort so that adapting the material to
individual learners' language competence and interests is beyond the
reach of traditional methods. To overcome this limitation, we generate
activities based on the target text for which the user wants to
acquire the vocabulary. While there are multiple activity types one
could consider, we implemented multi-gap activities
\cite{zesch-melamud-2014-automatic}, given that they make it possible
for the learner to engage with a word using several sentence contexts
drawn from the targeted book. Given the frequency threshold used in
constructing the domain model graph, there are at least five sentence
for each word in the book. We rank sentences to determine which
sentences are best out of context.  Fortunately this issue has been
addressed in lexicography, where authentic sentences are used in
dictionaries to illustrate word usage. \newcite{kilgarriff2008gdex}
developed GDEX, a method to identify sentences that are well-suited to
illustrate word meaning within a single sentence context. GDEX
considers factors such as the sentence length, use of rare words and
anaphora, target word occurrence in main clauses, sentence
completeness, and target word collocations towards the end of
sentences.  Sentence length and rare word usage are the highly
weighted features. We adapted GDEX for our purpose of ranking
sentences for vocabulary learning activities and customized the rare
word feature to reflect the individual learner's vocabulary knowledge
as recorded in the learner model.

\begin{figure}[htb]
\centering
\includegraphics[width=.7\textwidth, trim={0.3cm 3.9cm .6cm 2.1cm},clip = true]{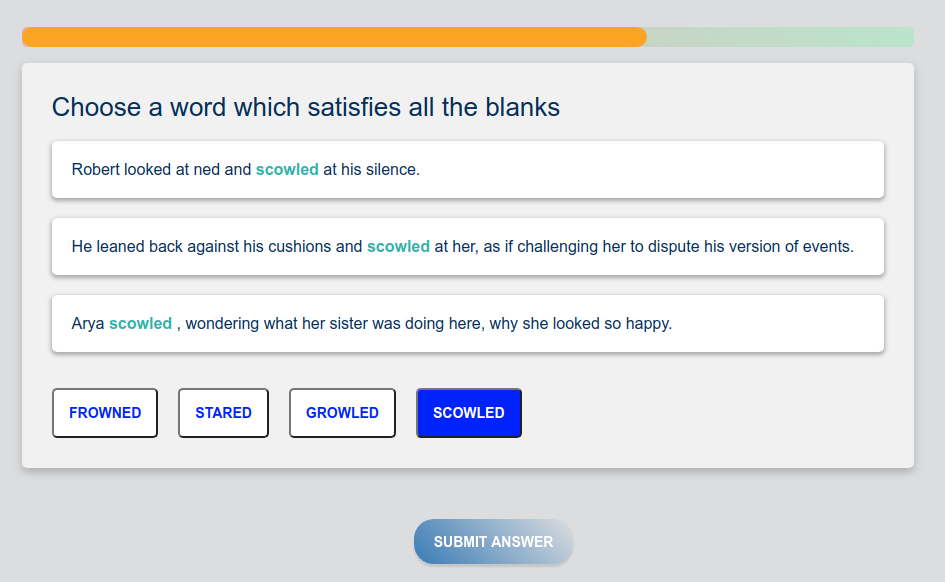}
\caption{An example activity}
\label{fig:activity}
\end{figure}

Learning and testing in the system is conducted in sessions. Each
session consists of the top 20 central nodes from the learner model
that are below the mastery score threshold.  Multi-gap activities
consisting of three to four sentences in which the target word chosen
from the central node word family occurs are used for both learning
and testing. The sentences are initially shown with the occurrences
replaced by a blank. For each activity four lexical options are
provided, the target word and three distractors chosen from the
vocabulary space of the book as discussed below.
Figure~\ref{fig:activity} shows an example activity targeting
the word family \emph{scowl} in a learning session for the book "A
Game of Thrones: A Song of Ice and Fire", after the correct word was
selected by the learner.

In the learning mode, the learners are provided with learning aid such
as dictionary look-up, translations, and word usage examples from
within and outside the targeted book.  The mastery scores in the
learner model are not updated during training mode. In the testing
mode, no such support is provided and the master score for the target
family and its neighbours are updated based on the user responses.

Distractor generation is a critical part of multi-gap learning
activities. If the distractors are randomly selected or heterogeneous
in linguistic characteristics such as POS, this can make it easy to
choose the right option for the given sentence contexts. We instead
are interested in distractors that require more cognitive effort to
discriminate, actively engaging the learner with the different
sentence contexts. We base the distractor selection on the learner
model. As this graph is built to encode semantically-related words
near each other, many of the immediate neighbours are (near) synonyms so that
there is a substantial danger of them fitting the gaps. By
experimenting with different node distances, we found that choosing
second neighbours as distractors that have the same POS as the target
word works best. We are also considering combining the distractor
generation based on the learner model with other distractor generation
strategies discussed by \citet{zesch-melamud-2014-automatic}.

\section{Conceptual Evaluation and Related Work}
\label{sec:evaluation-and-related-work}


Given the nature of the approach we presented, the evaluation and
comparison with related work has to evaluate conceptual advances and
put them into context. Our approach can be characterized by the
following aspects: First, the user can select what they want to learn
the vocabulary for; they pick the text of the book they want to be
able to read, i.e., the functional task goal. Second, the system
automatically creates a domain model graph representing the lexical
semantic space to be learned.  Third, a learner model is created as an
overlay of the domain model graph and records the mastery of the
concepts by the learner, with updates to the learner model spreading
activation through the graph to indirectly activate related concepts
as a way to avoid explicit interaction for every word.  Fourth, it
determines in which order the words can be learned in such a way that
the lexical space is efficiently explored, prioritizing the words that
are central nodes.  Fifth, the system compactly represents word
families to allow the visualization and open learner model to be
concise and usable with minimal number of interactions. Sixth, the
system supports learning of the words using multi-gap activities using
sentences drawn from the authentic context of the book to be learned.

Putting this approach into context of the related work on vocabulary
learning, there is a large number of applications designed to support
vocabulary learning -- though, as we will see, the above
characteristics clearly seem to set our approach apart from what is
offered in this domain.

Foreign language textbooks systematically provide a list of vocabulary
items per chapter and there are many specialized or general file card
applications for memorizing these sentences including
\emph{Phase-6.de}, \emph{Quizlet.com}, or \emph{Ankiweb.net}.  Other
applications offer more language-related functionality.

\emph{Lextutor} (\url{https://lextutor.ca}) is a website offering a
collection of tools to learn vocabulary using lexical resources such
as frequency-based vocabulary lists and corpus data.  \emph{List
  learn} supports learners in choosing words from frequency-based word
lists and work with corpus concordances.  \emph{Grouplex} lets the
learner select from a 2k crowd-sourced word list and practice them in
fill-in-the-blank activities, with hints based on dictionary
definition and POS tags.  \emph{Flash} employs cards showing words on
one side and lexical support on the other. Apart from word meaning and
usage, \emph{MorphoLex} supports learning regular inflectional and
derivational affixes based on the word family levels of
\newcite{bauer1993word}.  Other lextutor tools target reading texts
with support from concordances and dictionaries. \emph{Resource
  assisted reading} lets the user choose a pre-processed book, but
\emph{Hyper text} allows the learner to upload their own text. While
lextutor offers a variety of tools and corpus resources, none of them
offer personalized learning, performance tracking, or structured
vocabulary spaces.

\emph{Memrise.com} is a flashcard based de-contextualized commercial
vocabulary learning application focused on beginners, with learning
units grouped by theme with little freedom for the learner to choose
contents of interest. \emph{Duolingo.com} is a strictly guided
application supporting the users to learn foreign language using
various learning activities offering some gamification elements but no
personalized vocabulary learning for texts or domains of personal
interest. \emph{Vocabulary.com} is a gamified free vocabulary list
learning application that lets learners choose from collections and
the literature to practice the words in multiple choice questions
activities to choose the correct meaning phrase for the given word
usage. The literature only is a source of vocabulary though, it is not
used as testing context or learning goal, and the vocabulary domain is
not semantically structured or to construct a structured learner
model.
\emph{Cabuu.app} supports
learning of vocabulary lists scanned from books by associating each
item with gestures.

Overall, while there is a rich landscape of applications supporting
vocabulary learning, the six characteristics of the method presented
in this paper set our approach apart -- especially the use of
distributional semantic methods to create a graph representation for
any book or text the user wants to read, to efficiently organize and
individually support and track the learning in this lexical space.

\section{Conclusion}
\label{sec:conclusion}

In this paper, we discussed the methodological basis and realization
of a tool allowing the learner to systematically learn the lexical
material needed to be able to read a book they are interested in.
Automatically structuring the lexical space and sequencing the
learning is achieved through distributional semantic methods, the
automatic identification of word families, and concepts from network
analysis. The graph-based domain model that is automatically derived
from the given book serves as the foundation of a learner model
supporting the selection of an efficient learning path through the
lexical space to be acquired. Multi-gap activities are automatically
generated from the targeted book and used for practice and testing
activities. 


In addition to self-guided learning for people interested in reading
specific books, which may be particularly useful in the context of
so-called intensive reading programs, the approach is particularly
well-suited for the English for Specific Purposes context, where both
the language and the particular content domain are of direct
importance.  Given this kind of integration of language and content
learning, a similar affinity exists to so-called Content and Language
Integrated Learning \cite{coyle2010content}. 






\bibliographystyle{vocab}
\bibliography{vocab}
\end{document}